\documentclass[10pt,twocolumn,letterpaper]{article}

\usepackage[pagenumbers]{cvpr} 

\usepackage{graphicx}
\usepackage{amsmath}
\usepackage{amssymb}
\usepackage{booktabs}
\usepackage{multirow}

\usepackage[pagebackref,breaklinks,colorlinks]{hyperref}
\usepackage{cuted}
\usepackage{capt-of}

\usepackage[accsupp]{axessibility}
\usepackage[capitalize]{cleveref}
\crefname{section}{Sec.}{Secs.}
\Crefname{section}{Section}{Sections}
\Crefname{table}{Table}{Tables}
\crefname{table}{Tab.}{Tabs.}

\def\cvprPaperID{816} 
\def\confName{CVPR}
\def\confYear{2022}

\begin{document}
\title{What's in your hands? 3D Reconstruction of Generic Objects in Hands}
\author{Yufei Ye\textsuperscript{12} \qquad Abhinav Gupta\textsuperscript{12} \qquad Shubham Tulsiani\textsuperscript{1}   \\
\textsuperscript{1}Carnegie Mellon University  \qquad \textsuperscript{2}Meta AI \\
{\tt \small yufeiy2@cs.cmu.edu \qquad gabhinav@fb.com \qquad shubhtuls@cmu.edu} \\
{\tt \small \href{https://judyye.github.io/ihoi/}{https://judyye.github.io/ihoi/}}
}

\maketitle

\begin{strip}\centering
\vspace{-1.7cm}
\includegraphics[width=\textwidth]{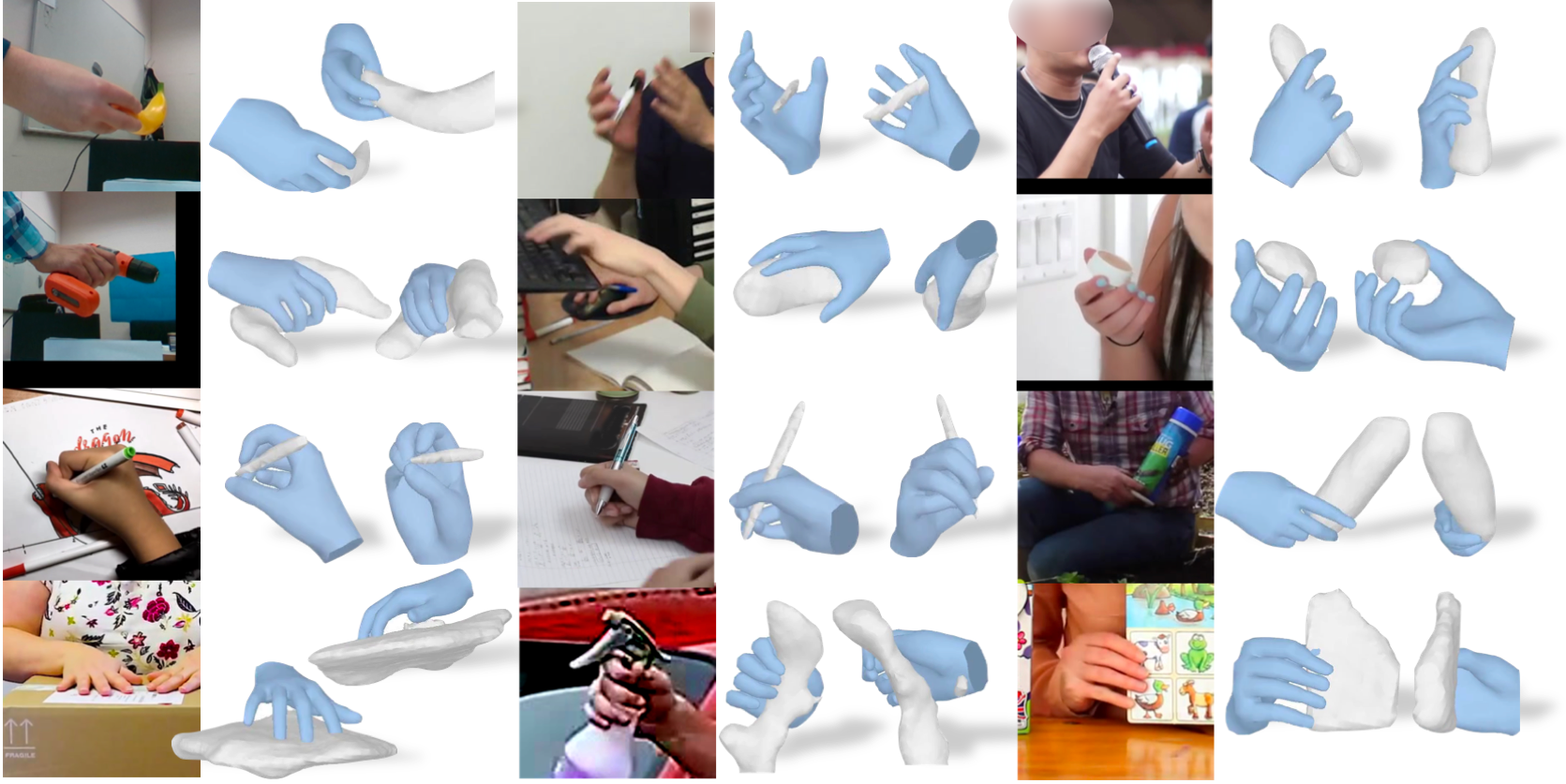}
\captionof{figure}{Given an RGB image depicting a hand holding an object, we infer the 3D shape of the hand-held object (rendered in the image frame and from a novel view).  
\label{fig:teaser}}
\vspace{-.2cm}
\end{strip}

\begin{abstract}
   Our work aims to reconstruct hand-held objects  given a single RGB image. In contrast to prior works that typically assume known 3D templates and reduce the problem to 3D pose estimation, our work reconstructs  generic hand-held object without knowing their 3D templates. Our key insight is that hand articulation is highly predictive of the object shape, and we propose an approach that conditionally reconstructs the object based on the articulation and the visual input.  Given an image depicting a hand-held object, we first use off-the-shelf systems to estimate the underlying hand pose and then infer the object shape in a normalized hand-centric coordinate frame. We parameterized the object by signed distance which are inferred by an implicit network which leverages the information from both visual feature and  articulation-aware coordinates to process a query point.   We perform experiments across three datasets and show that our method consistently outperforms baselines and is able to reconstruct a diverse set of  objects.
   We analyze the benefits and robustness of explicit articulation conditioning and also show that this allows the hand pose estimation to further improve in test-time optimization.
\end{abstract}
\vspace{-3mm}

\vspace{-1em}
\section{Introduction}
We humans use our hands to affect the world around us. From turning off the alarm clock in the morning to  cleaning dishes after dinner, we continually interact with generic objects while accomplishing diverse tasks. An intelligent agent aiming to mimic human interactions or a virtual assistant striving to aid in them must be able to understand such generic everyday interactions from perceptual input. We argue that this understanding should go beyond assigning semantic labels (`touching', `holding') and instead model the physical nature of the observations. Towards this, we pursue a \emph{geometric} representation of hand-object interactions -- given a single image depicting a human hand interacting with a generic object, we aim to infer the 3D shape of the hand-held object.

Over the past decade, we have made significant advances in inferring the 3D shape of both hands and objects in isolation. Hand poses can be recovered in the form of 2D keypoints, 3D skeletons \cite{mueller2018ganerated ,mueller2019real,panteleris2018using,iqbal2018hand,zimmermann2017learning}, or even full 3D meshes \cite{boukhayma20193d,rong2020frankmocap} via either fitting templates or direct regression. On the other hand, recent works have also pursued scaling object reconstruction from estimating the 6D pose of one specific known object \cite{roberts1963machine} to more generic objects, such as various instances within one category  \cite{kanazawa2018learning,li2020self,goel2020shape}, or even pursuing a joint model for cross-category reconstruction \cite{gkioxari2019mesh,choy20163d}. But one area where the progress has been quite limited is understanding human-object interactions  (HOI) specifically for manipulable objects~\cite{shan2020understanding,Damen2018ScalingEV}.

Reconstruction of objects in hand in-the-wild is highly challenging and ill-posed due to lack of data, presence of mutual and self-occlusion. Current works~\cite{Hasson2020LeveragingPC,tekin2019h+,liu2021semi,garcia2018first,Cao2020ReconstructingHI} typically focus on reconstructing a objects with known templates, thus reducing the task to 6D pose estimation. We argue that knowing the 3D template of the object as a priori during inference is a strong assumption and prevents these systems from reconstructing unknown objects. Furthermore, they struggle to handle various object shapes in the wild as these templates are rigid and instance-specific.  In contrast, our work studies hand-object reconstruction without object templates and instead focuses on reconstructing HOI for novel objects from images.

Our key observation is that hand articulation is driven by the local geometry of the object. Thus, hand articulation provides strong cues for the object in interaction. Fingers curled like fists indicate thin handles in between while open palms are likely to interact with flat surfaces. Instead of treating the hand occlusion as noise to marginalize over, we explicitly consider hand pose as informative cues for the object it interacts with. We operationalize this idea by conditionally predicting the object shape based on hand articulation and the input image. Instead of estimating both hand pose and object shape jointly, we leverage advances in hand pose reconstruction to estimate hand pose first. Given the inferred articulated hand along with the input image, our approach then reconstructs the object in a normalized hand-centric coordinate frame. 

We evaluate our method across three datasets including synthetic and real-world benchmarks and compare ours with prior explicit and implicit HOI reconstruction methods that infer the shape of unknown objects independent of hand pose. Our articulation-conditioned object shape prediction consistently outperforms prior works by large margins and can reconstruct various objects in a wide range of shapes. We also analyze how our model benefits from articulation-aware coordinates. Lastly, we show that the initial hand pose estimation could be further improved by encouraging interaction between the predicted hand and the object.   

\section{Related Work}


\noindent\textbf{Hand pose estimation. } Approaches tackling hand pose estimation from RGB(-D) images can be broadly categorized as being model-free and model-based. Model-free methods~\cite{mueller2018ganerated,choi2020pose2mesh,rogez20143d,rogez2015understanding,mueller2019real,panteleris2018using,iqbal2018hand,zimmermann2017learning} typically detect 2D keypoints and lift them to 3D joints position or hand skeletons. 
Some works~\cite{choi2020pose2mesh,pemasiri2021im2mesh,ge20193d} then directly predict 3D meshes vertices from the 3D skeleton by coarse-to-fine generation.  Model-based methods\cite{rong2020frankmocap,zhang2019end,sridhar2015fast, boukhayma20193d,zhou2020monocular} leverage statistical models like MANO~\cite{romero2017embodied} whose low-dimensional pose and shape parameters can be directly regressed~\cite{boukhayma20193d,rong2020frankmocap} or optimized ~\cite{zhang2019end,zhou2020monocular,sridhar2015fast}. 
These model-based methods are generally robust to occlusion, domain gap \etc, and we build on these in our work. In particular, we use a state-of-the-art model-based reconstruction system~\cite{rong2020frankmocap} to first estimate hand pose and condition the inference of the hand-held object shape on this prediction. While our work primarily focuses on inferring the object shape given an off-the-shelf hand pose estimate, we also show that jointly reasoning about the geometric interaction between the predicted 3D object and the inferred hand pose can help improve the initial hand pose estimate. 

\noindent\textbf{Single-view Object Reconstruction.} Reconstructing objects from images is a long-standing problem, dating back to the seminal thesis from Larry Roberts~\cite{roberts1963machine}, where 3D models were known and the problem was reduced to 6-DoF pose estimation. In the following decades, several works have since aimed to reconstruct more generic objects from images~\cite{agarwal2011building,hoiem2005automatic,gupta2010blocks,kundu20183d}. Recent learning based-methods can learn category-specific networks for 3D prediction~\cite{kar2015category, cashman2012shape,cootes1992active} across a broad class of objects, and can even do so without direct 3D supervision~\cite{kanazawa2018learning,goel2020shape,li2020self,ye2021shelf}.  Using stronger 3D supervision, other approaches have shown that it is possible to learn a common model across multiple categories, with output representations such as voxels~\cite{girdhar2016learning,wu2016learning,choy20163d}, meshes~\cite{gkioxari2019mesh,groueix2018papier,wang2018pixel2mesh}, point clouds~\cite{fan2017point,lin2018learning}, or primitives~\cite{deng2020cvxnet,genova2020local,tulsiani2017learning}.  Closer to our work, neural implicit representations \cite{park2019deepsdf,chen2019learning,mescheder2019occupancy} have shown the impressive capacity for accurate reconstruction for different topology, and our work extends these to be articulation-conditioned for inferring 3D of hand-held objects. While the field of object reconstruction has witnessed remarkable progress, the state-of-the-art methods typically assume isolated and unoccluded objects in images -- and cannot be directly leveraged for reconstructing hand-held objects.  Even approaches that are robust to occlusion consider it as noisy context to marginalize over, instead of a source of signal for the shape of the underlying object. In contrast, our work shows that explicitly taking hand pose into account helps infer the 3D structure of objects more accurately.  



\begin{figure*}
    \centering
    \includegraphics[width=\linewidth]{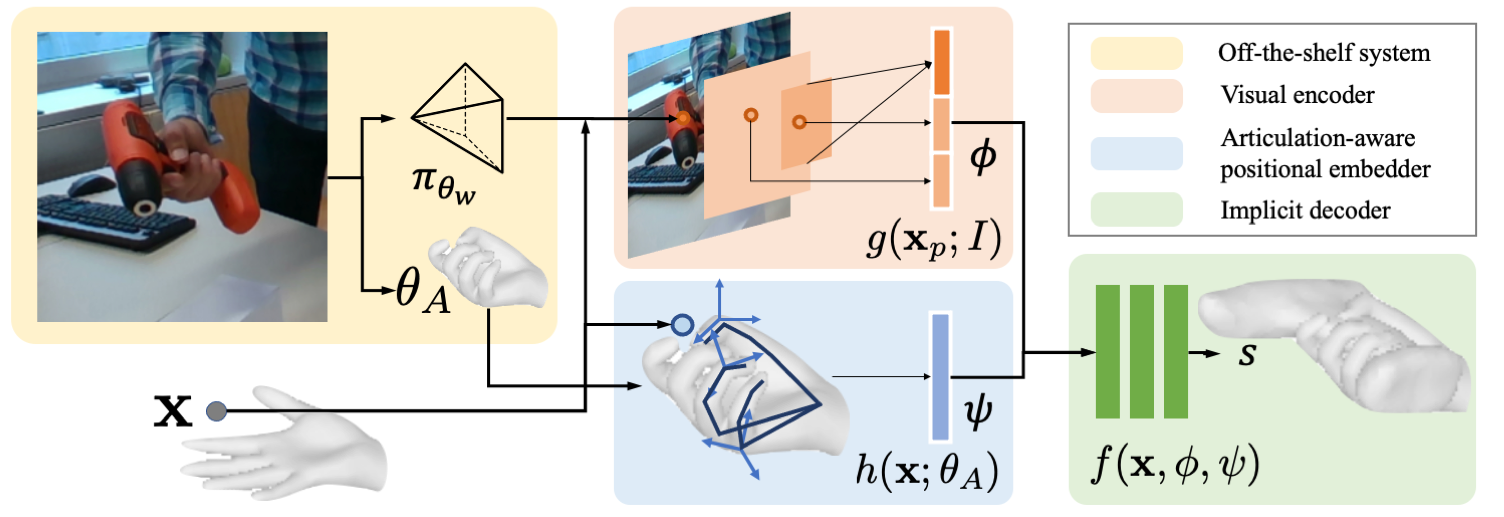}
    \caption{Given an image of a hand-held object, we first use an off-the-shelf system to estimate hand articulation $\theta_A$ and the camera pose $\pi_w$. With the predicted articulated hand along with the image, the object shape is reconstructed by an implicit network. For each query point $\mathbf x$ in canonical hand wrist frame, it is transformed to image space $\mathbf x_p$ to get visual feature $\phi=g(\mathbf x_p, I)$. In parallel, we also encode its articulation-aware representation $\psi=h(\mathbf x;\theta)$.  Then we use an implicit decoder to predict signed distance value $s=f(\mathbf x, \phi, \psi)$.}
    \label{fig:method}
\end{figure*}

\vspace{1mm}
\noindent \textbf{Reconstructing hand-held objects. } Understanding HOI in 3D is a very challenging problem due to mutual occlusion and lack of data with annotation. To make inference tractable, prior works typically make the simplifying assumption of having access to known instance-specific templates (typically around 10 objects) \cite{tekin2019h+,hamer2010object,hamer2010object,romero2010hands,tzionas2016capturing,sridhar2016real,tzionas20153d,garcia2018first,hampalihonnotate} and infer these known objects under lab controlled environments.
With access to instance-specific templates, they reduce the object reconstruction to 6D pose estimation and jointly predict both hand and object pose by reasoning about their interactions. The joint reasoning could be  implicit feature fusion~\cite{liu2021semi,shan2020understanding,Chen2021JointH3,tekin2019h+, Gkioxari2018DetectingAR}, explicitly leveraging geometric constraints like contact or collision~\cite{Grady2021ContactOptOC,Brahmbhatt2020ContactPoseAD,Cao2020ReconstructingHI,Zhang2020Perceiving3H,Corona2020GanHandPH,hamer2010object}, or encouraging physical realism~\cite{pham2017hand,tzionas2016capturing}.  In contrast, we focus on inferring hand-held object shape without knowing a corresponding template. Sharing this goal, work by Hasson \etal \cite{hasson19_obman} predicts explicit genus-0 meshes and Karunratanakul \etal \cite{karunratanakul2020grasping} predict a joint implicit field for reconstructing the hand and object. While these methods also perform model-free reconstruction, unlike our approach they \emph{independently} infer the object shape and hand pose in a feed-forward manner. Instead,  we formulate object reconstruction as conditional inference of 3D shape given the hand-articulation, and make our predictions in a normalized wrist frame, and show that this significantly improves performance.

\section{Method}
Given an image depicting a hand holding an object, we aim to reconstruct the 3D shape of the underlying object.
Our key insight is that the hand articulation is predictive of the object shape within it, for example, fingers pinching together indicate a thin stick-like structure between them.
We operationalize this by explicitly conditioning the inference of object shape on the (predicted) hand articulation.

As shown in Fig \ref{fig:method}, we first use an off-the-shelf system to estimate hand articulation and predict the camera transformation that projects the canonical articulated hand to the image coordinates. Given the predicted hand along with the input image, we then infer the object shape via an articulation-conditioned reconstruction network. This network is implemented as a point-wise implicit function~\cite{park2019deepsdf} that maps a query 3D point to a signed distance from the object surface, and the zero-level set of this function can be extracted as the object surface~\cite{lorensen1987marching}. Instead of predicting this 3D shape in the image coordinate frame, our implicit reconstruction network infers it in a normalized frame around the hand wrist. This allows the network to learn relations between the hand articulation and object shape that are invariant to global transformations.

More formally, given an input image $I$, we first infer the underlying the hand pose $\theta$ and the camera pose $\pi$. Then, for any point $\mathbf x$ in the normalized wrist frame, the object inference model takes in the query point with the image and predicts its signed distance function $s$. More specifically, the projection of the point to image coordinates is used to obtain corresponding visual features $\phi = g(\pi(\mathbf x); I)$. In parallel, we  also encode its position relative to each hand joint to extract an articulation-aware representation $\psi = h(\mathbf x; \theta)$. The point-wise visual feature and articulation embedding are then used by an implicit decoder to predict signed distance value $s =  f(\phi, \psi)$ at the query point $\mathbf{x}$. 




\subsection{Preliminary: Hand Reconstruction}

We use an off-the-shelf system \cite{rong2020frankmocap} to estimate hand articulation and associated camera pose from an input image. The reconstruction method is model-based which directly regresses a 45-dimensional articulation parameter ($\theta_A$) and a 6-dim global rotation and translation ($\theta_w$) along with a weak perspective camera. We rig the parametric MANO model by the predicted articulation pose $\theta_A$ to obtain an articulated hand mesh in a canonical frame around the wrist. 
To relate a point in the wrist frame to the image space, we first transform the hand by the predicted global transformation and then project it by the camera matrix. As an implementation detail, we convert the predicted weak perspective camera to a full perspective one as it helps to account for large perspective effects.  In summary, we project a query point in the canonical wrist frame to the image by 
$$
\mathbf x_p = \pi_{\theta_w}(\mathbf x) = KT_{\theta_w}\mathbf x
$$ 
where $K$ is the camera intrinsic and $T_{\theta_w}$ is the global rigid transformation of the hand.

\subsection{Articulation-conditioned SDF}
Given the predicted hand articulation $\theta_A$, and camera matrix $\pi_{\theta_w}$, our articulation-conditioned object shape inference network takes an additional input image $I$ to output a signed distance field of the object.  For a query point in the wrist coordinate frame, the point-wise network takes into account the query's corresponding visual feature from a visual encoder and its relative position to each joint from an articulation-aware positional embedder.
The visual feature and the embedding are then passed to an implicit decoder along with the query to predict the signed distance. 

\noindent\textbf{Visual encoder.} The visual encoder first extracts the image feature pyramid at different resolutions.  For a query 3D point in the wrist frame, the visual encoder projects it to the image coordinate and compute global and local feature from the pyramid. The global part allows us to reason about global context and generate more coherent object shapes. For example, realizing the object is a bottle helps the network to generate a cylinder shape. The local feature allows the prediction more consistent with the visual observation~\cite{Yu2021pixelNeRFNR,Saito2019PIFuPI}. 

The backbone of the visual encoder is implemented as ResNet~\cite{He2016DeepRL}. The global feature is a linear combination of the averaged conv5 feature. The local feature for each point is an interpolated feature at image coordinate from where it is projected by the predicted camera $\phi[\pi_{\theta_{w}}(\mathbf x)]$, where $\phi$ denotes resnet feature and $\phi[x]$ represents a bilinear sample of the feature at a 2D location $x$. The local feature sampling is implemented for every layer of the feature pyramid $\phi_{1,\dots,5}[x]$. It allows the model to draw visual cues with various resolutions and receptive fields.

\noindent \textbf{Articulation positional embedder. } 
Our key idea is that the hand pose is predictive of the object shape it interacts with. This is especially informative and complements the visual cues for reconstructing hand-held objects that are often occluded. We explicitly encode hand pose information for the query point via the articulation embedder. 
To do this,
one naive way is to simply use identity mapping  $\psi = \theta_A$. However, this representation is not robust to hand prediction error as we show in ablation. Furthermore, it is not trivial for the network to relate the reconstruction metric space with the hand pose joint space. For example, if a point is within 2mm from both index and thumb, it is very likely to have some object passing through. To better capture the structure of the problem, we encode the hand pose information by the position of the query points  relative to the articulated joints. 




More specifically, the articulation embedder takes as input an articulation parameter $\theta_A$ and a point position in wrist frame $X$ to output the articulation-aware encoding $\psi = h(X;\theta_A)$. 
The encoding is a concatenation of the coordinates relative to every joint. Given the articulation parameter $\theta_A$, we run forward kinematics to derive transformation $T(\theta_A): \mathbb R ^3 \to \mathbb R ^ {45}$ that maps a point in wrist frame to each joint coordinate. 
The 15 joint coordinates are position encoded \cite{Vaswani2017AttentionIA} and concatenated together as the final representation $h(\mathbf x; \theta_A) = \gamma(T(\theta_A)\mathbf x) $ where $\gamma$ is the positional encoder. For more details please refer to appendix. 


\noindent\textbf{Implicit decoder. }
The decoder maps the query points with visual feature and articulation embedding to a signed distance value $f(\mathbf x, \phi, \psi) = s$. These two representations $\phi, \psi$ are concatenated together and passed along with the query point to the decoder.  The decoder simply follows the design in DeepSDF \cite{park2019deepsdf} which consists of 8 layers of MLP with a skip layer. 

\subsection{Training}
To learn our articulation conditioned neural implicit field, we rely on a training dataset where we assume known hand pose and 3D shape of the object in the image frame. 
We preprocess the data by sampling   points inside and outside of the object around the hand to calculate the ground-truth  SDF. 95\% of the points are sampled around the surface of the objects and others are sampled uniformly in the space. During training, the network optimizes to match the predicted SDF to the ground-truth at the sampled points with the eikonal term as a regularizer. 

$$\mathcal L = \|s - \hat s\| + \lambda (\|\nabla s\| - 1 )^2$$

After the network is learned, at inference time, we do not require knoledge of the object 3D shape in the canonical frame  a priori which is a major limitation of most  most prior works.



\subsection{Refining hand pose}

While our work primarily focuses on object reconstruction conditioned on a predicted hand pose, this initial pose prediction, while reliable, is not perfect. As object reconstruction also leverages visual cues, our insight is that it can provide complementary information to further refine the predicted hand pose. During inference, we show that the predicted hand pose and object shape can therefore be further (jointly) optimized by enforcing physical plausibility -- by encouraging contact while discouraging intersection. 

We optimize the articulation pose parameters with respect to these two interaction terms, which can be naturally incorporated with an SDF representation. To discourage intersection between the hand and object, we penalize if the points on the hand surface are predicted to have negative SDF values by the object reconstruction model. Following prior work~\cite{hasson19_obman}, we encourage hand-object contact for specifically defined regions -- if the surface points in these contact regions are near the object surface, they are encouraged to come even closer.

\begin{align*}
\min_{\theta} &\sum_{\mathbf x\in \mathcal H}\|\max(-f(\mathbf x), 0) \|  + \\
& \sum_{\mathbf x\in \mathcal C} \max(\| \min(f(\mathbf x) - \tau, 0) \| - \epsilon, 0)
\end{align*}

Note that the SDF $f$ is conditioned on articulation thus it is also a function of the hand pose $\theta$. As we refine the hand pose, the SDF of the object also changes accordingly. One could continuously update the SDF every time the pose is updated but we use a simpler solution that fixes the SDF during hand pose optimization and only update it once using the final optimized pose $\theta^{\*}$. 



\section{Experiments}

\begin{table*}[]
\small
    \centering
    \vspace{-2mm}
    \begin{tabular}{l   cccc | cccc| cccc}
    \toprule
    & \multicolumn{4}{c}{ObMan} & \multicolumn{4}{c}{HO3D} &  \multicolumn{4}{c}{MOW} \\ 
    & F-5 $\uparrow$ & F-10 $\uparrow$  & CD $\downarrow$  & Vol $\downarrow$ 
    &  F-5 $\uparrow$ & F-10 $\uparrow$  & CD $\downarrow$  & Vol $\downarrow$  
    & F-5 $\uparrow$ & F-10 $\uparrow$  & CD $\downarrow$ & Vol $\downarrow$  \\
    \midrule
        HO \cite{hasson19_obman}  & 0.23 & 0.56 & \textbf{0.64} & 8.64 
        & 0.11	& 0.22	& 4.19  & 9.44
        & 0.03	& 0.06 & 49.8  & 25.6 \\
        GF \cite{karunratanakul2020grasping} & 0.30&	0.51 & 1.39 &  1.84
        & 	0.12	& 0.24 &	4.96 & 6.31
        & 	0.06	& 0.13	& 40.1 & \textbf{8.82} \\
    Ours & \textbf{0.42} & \textbf{0.63} & 1.02 &  \textbf{1.74 }
    &  \textbf{0.28}	& \textbf{0.50} &\textbf{	1.53} & \textbf{4.77}
    &  \textbf{0.13} & \textbf{0.24} &	\textbf{23.1} & 19.4 \\
    \bottomrule
    \end{tabular}
    \vspace{-2mm}
    \caption{Quantitative results for  object reconstruction error using F-score ($5mm,10mm$), Chamfer distane ($mm$) and intersection volume ($cm^3$).  We  compare our method with prior works \cite{hasson19_obman, karunratanakul2020grasping} on Obman, HO3D, MOW datasets.}
    \vspace{-1em}
    \label{tab:main}
\end{table*}

\begin{figure}
    \centering
    \includegraphics[width=\linewidth]{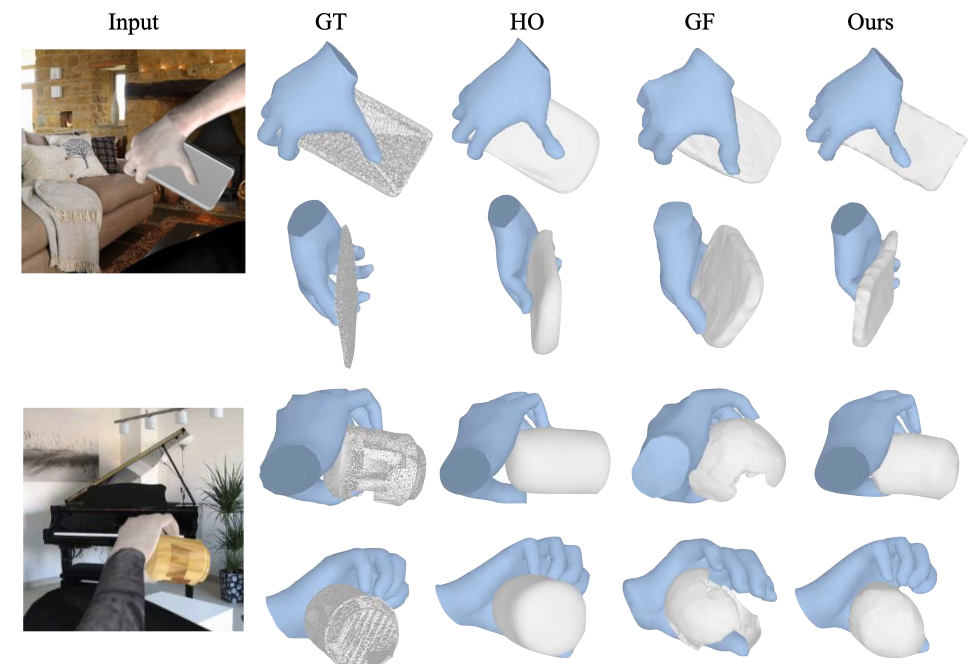}
    \caption{Visualizing reconstruction from our method and two baselines\cite{hasson19_obman, karunratanakul2020grasping} on ObMan dataset from the image frame and a novel view.}
    \label{fig:main_fig}
\vspace{-2em}
\end{figure}

We compare our method with two model-free baselines\cite{hasson19_obman,karunratanakul2020grasping} on three datasets -- one synthetic, and two real-world. We show that our approach outperforms baselines across these datasets, both in terms of object reconstruction and modeling hand-object interaction. We further analyze the benefit from explicitly considering hand pose and the benefit from our particular form of articulation-aware positional encoding. Lastly, we show that our reconstructed hand-held object could further refine the initial hand pose estimation and improve hand-object interaction.

\vspace{1mm}
\noindent\textbf{Datasets and Setup. } We evaluate our method across three datasets.
\begin{itemize}
    \item ObMan~\cite{hasson19_obman} is a synthetic dataset that consists as 8 categories of 2772 objects from 3D warehouse~\cite{Chang2015ShapeNetAI}. The grasps are automatically generated by GraspIt \cite{miller2004graspit}, resulting in a total of 21K grasps. The grasped objects are rendered over random backgrounds using Blender. We follow the standard splits  where there is no overlap between the objects used in training and testing. 
    
    \item HO-3D\cite{hampali2020honnotate} is a real-world video dataset consisting of 103k annotated images capturing 10 subjects interacting with 10 common YCB objects~\cite{alli2015TheYO}. The ground- truth is annotated using multi-camera reconstruction pipelines. 
    To test on more shapes, we create a custom split by holding out one video sequence per object as test set. Please refer to the appendix for more details.

    \item MOW \cite{cao2021reconstructing} dataset consists of a curated set of 442 images, spanning 121 object templates, collected from in-the-wild hand-object interaction datasets \cite{shan2020understanding,Damen2018ScalingEV}. It is  more diverse in terms of both appearance and object shape compared to the HO3D dataset, but only provides approximate ground-truth. These object shape and hand pose annotations are obtained via a single-frame optimization-based method \cite{cao2021reconstructing}. 
    We use 350 randomly selected examples for training and the remaining 92 for evaluation. 
\end{itemize}

For the ObMan dataset, we use the hand pose predictor from Hasson \etal \cite{hasson19_obman} as this system is specifically trained on this synthetic dataset. For  HOI and MOW, we use the FrankMocap~\cite{rong2020frankmocap} system that is trained on multiple real-world datasets. 
Since HOI data with ground truth in the real world is scarce and lacks diversity in terms of object shape and appearance, we initialize our method and baselines with models pretrained on ObMan and finetune them on HO3D and MOW datasets.

\noindent\textbf{Evaluation metrics. } We evaluate the quality of both, object reconstruction and the relation between object and hand.
To evaluate the reconstruction quality, we first extract a mesh from the predicted SDF. Following prior works, we then evaluate the object reconstruction by reporting Chamfer distance (CD), but also report the F-score at 5mm and 10mm thresholds as Chamfer distance is more vulnerable to outliers \cite{tatarchenko2019single}. Another desirable property for HOI reconstruction is that the interpenetration between the hand and the object should be minimal, we also report the intersection volume between two meshes (in $cm^3$) as a measure of understanding the relations between the hand and object.

\vspace{1mm}
\noindent\textbf{Baselines.} While most prior approaches require a known object template, recent work by Hasson \etal (HO)~ \cite{hasson19_obman} and Karunratanakul  \etal (GF)~\cite{karunratanakul2020grasping} can tackle the same task as ours -- inferring the shape of a generic object from a single interaction image. HO jointly regresses MANO parameters to estimate hand pose and reconstructs the object in the camera frame. It is based on Atlas-Net \cite{groueix2018papier}, and deforms a sphere to infer the object mesh. Closer to our approach, GF is also based on a point-wise implicit network. It takes an image as input and outputs an implicit field that maps a point in the camera frame to a signed distance to both, the hand and object, while also predicting hand part labels. 

Our approach differs from these baselines in three main aspects. First, both prior approaches infer object shape independent of the predicted hand pose, while we formulate hand-held object reconstruction as conditional inference. Second, these baselines encode the visual information only via a global feature while we additionally use pixel-aligned local features.
Third, while both baselines reconstruct objects in the camera frame, we predict them in a normalized  wrist frame with articulation-aware positional encoding. Note that while our approach predicts 3D in a hand-centric frame, the evaluations are all performed in the image coordinate frame for fair comparison (using the predicted hand pose to transform our prediction).


\begin{figure*}
    \centering
    \vspace{-.5em}
    \includegraphics[width=0.95\linewidth]{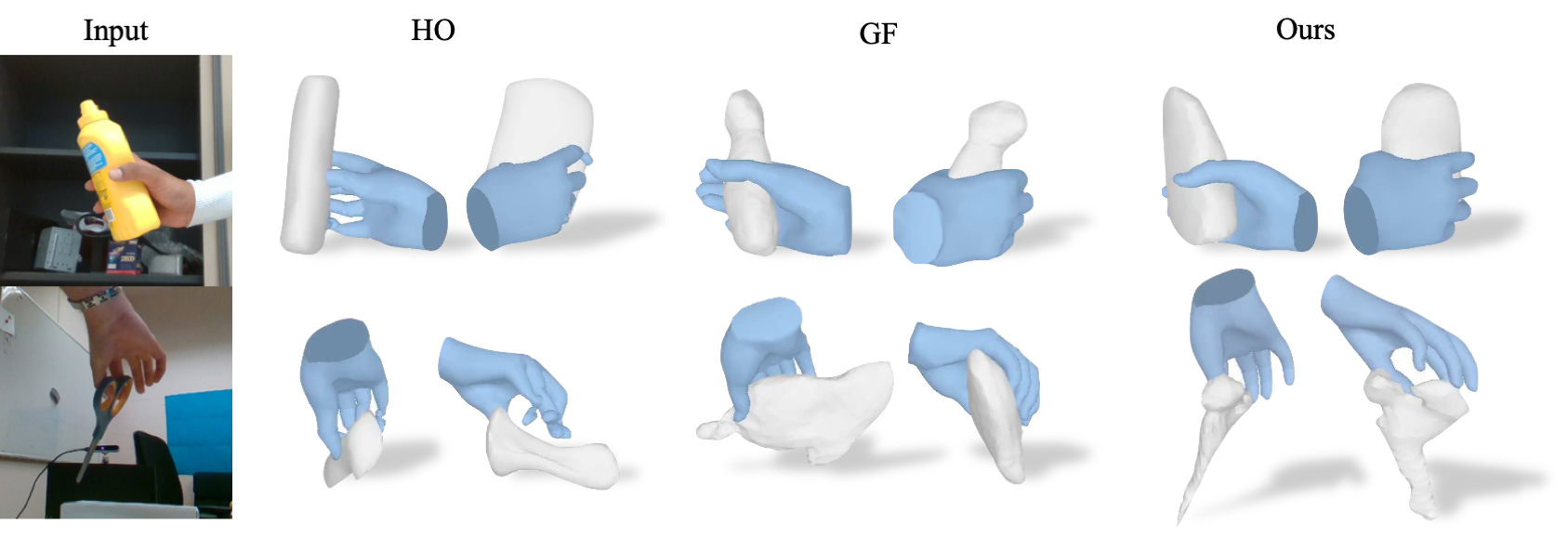}
\vspace{-1em}    \caption{
Visualizing reconstruction of our method and two baselines\cite{hasson19_obman, karunratanakul2020grasping} on HO3D dataset in the image frame and from a novel view.}
        \vspace{-1em}
\label{fig:ho3d}
\end{figure*}
\begin{figure*}
    \centering
    \includegraphics[width=0.95\linewidth]{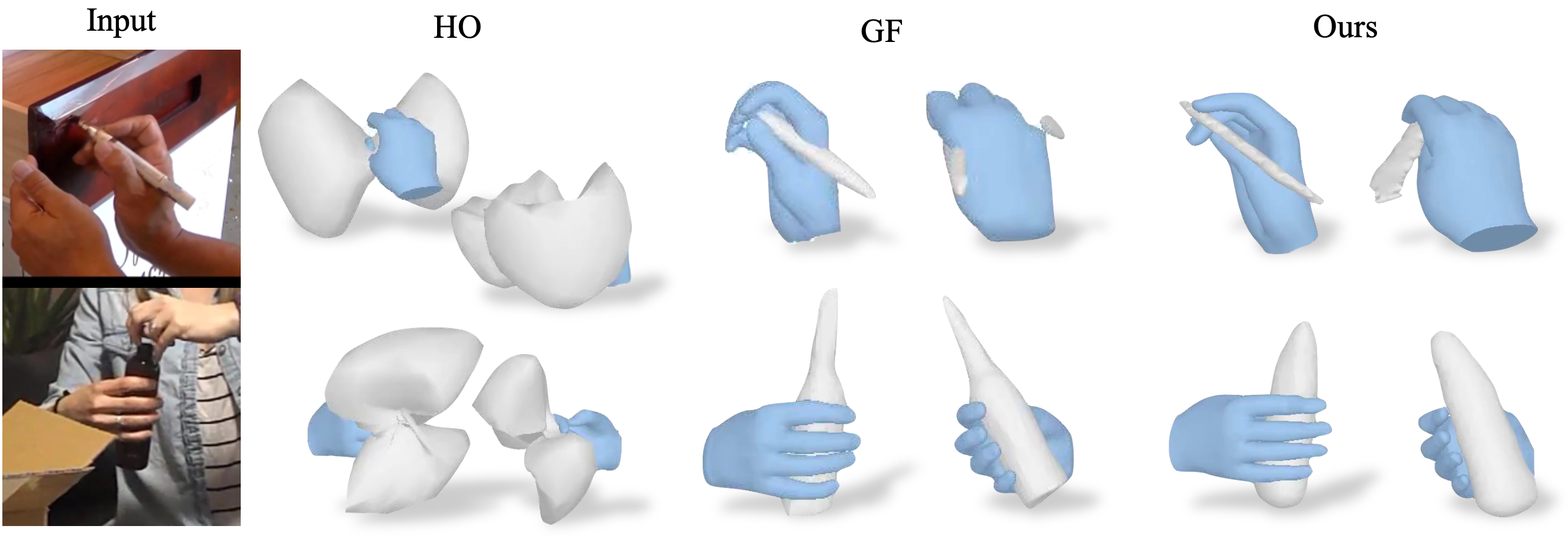}
    \vspace{-0.5em}
    \caption{Visualizing reconstruction of our method and two baselines\cite{hasson19_obman, karunratanakul2020grasping} on MOW dataset in the image frame and from a novel view.}
    \label{fig:rhoi}
    \vspace{-1em}
\end{figure*}

\subsection{Results on ObMan}

We visualize the reconstructed objects and the corresponding hand poses in Figure \ref{fig:main_fig}. While baselines can predict the coarse shape of the object, they typically lose sharp details such as the corner of the phone and sometimes miss part of the surface of the object. This may be because they only use a global feature of the image that loses spatial resolution. In contrast, our method reconstructs shape that better aligns with the visual inputs from the original view and hallucinate the invisible part of the objects occluded by hands.  

This is also empirically reflected in quantitative results reported in Table \ref{tab:main}.  We outperform baselines by a large margin on F-score. Our improvement over baselines is particularly significant on the smaller threshold, indicating that our method is better to reconstruct local shape. In terms of Chamfer distance, ours is better than GF that is also based on implicit fields. Ours is not as good as HO in Chamfer distance probably because HO explicitly trains to minimize Chamfer distance with a regularizer on edge length which discourages large displacements from a sphere thus producing fewer outliers. 



\subsection{Evaluation  on real-world datasets}

We visualize the reconstruction in the image frame and a novel view in Figure \ref{fig:ho3d} and Figure \ref{fig:rhoi}. 
GF can predict blobby cylinders but the reconstructed  objects lack details in shape such as around the neck of the mustard bottle, and sometimes reconstructs a different object shape such as predicting boxes instead of scissors. In contrast, our method is able to generate  diverse object shape more accurately including boxes, power drills, bottles, pens, cup, spray bottles etc.

\begin{table}[!htb]
    \footnotesize \centering
    \begin{tabular}{l   c c c}
    \toprule
   train set & F-5 $\uparrow$ & F-10 $\uparrow$  & CD $\downarrow$ \\
    \midrule
        ObMan & 	0.14 &	0.27 & 4.36 \\
        MOW  &\textbf{ 0.15}	& \textbf{0.30}	&\textbf{	4.09} \\
    \bottomrule
    \end{tabular}
    \caption{\textbf{Cross-dataset generalization: } we report quantitative results on HO3D for models pretrained on ObMan and MOW. }
    \vspace{-0.5em}
\end{table}

\noindent\textbf{Zero-shot transfer to HO3D. } We also directly evaluate  models that are only trained  on ObMan and MOW datasets and report their reconstruction results on HO3D dataset. Both models without finetuning still outperform baselines trained on HO3D dataset. 
Interestingly, even though the MOW dataset only consists of 350 training images, which is significantly less compared to 21K images from the synthetic dataset, learning from MOW still helps cross-dataset generalization. It indicates the importance of diversity for in-the-wild training. Please see the qualitative result in the appendix.

\subsection{Ablations} 

\begin{figure}
    \centering
    \includegraphics[width=\linewidth]{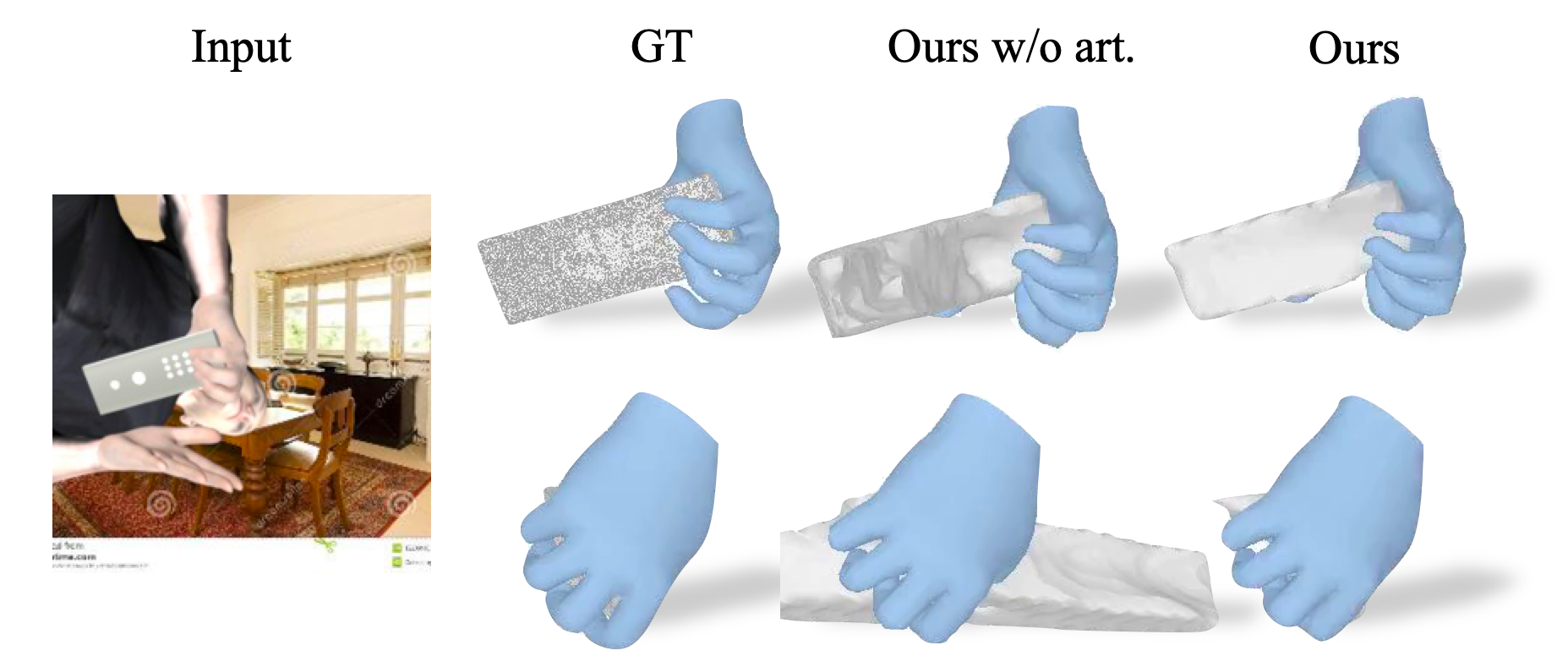}
    \caption{Visualizing reconstruction of hand-held object with or without explicitly considering  hand pose.  }
    \vspace{-3mm}
    \label{fig:wo_hand}
\end{figure}

\noindent \textbf{Importance of articulation conditioning. } We analyze how hand articulation conditioning helps hand-held object reconstruction by constructing a variant of our method that only conditions on pixel-aligned image features. This approach is analogous to the one proposed by Saito \etal~\cite{Saito2019PIFuPI} where human 3D shape is inferred by a pixel-aligned implicit network.  Table \ref{tab:wo_hand} reports results of this variant that do not explictily consider hand articulation and we observe that the object reconstruction degrades by a large margin while the intersection volume also doubles. This suggests that hand information provides a strong cue that is complementary to visual inputs. Figure \ref{fig:wo_hand} visualizes comparison between ours and the variant where our method can better respect hand-object physical relations such as objects do not penetrate the hands and the area around fingertips are likely to be in contact with objects.

\begin{table}
    \footnotesize \centering
    \begin{tabular}{l l   c c c c}
    \toprule
   &    & F-5 $\uparrow$ & F-10 $\uparrow$  & CD $\downarrow$ & Vol $\downarrow$\\
    \midrule
\multirow{2}{*}{ObMan}   &     Ours  & \textbf{0.42} & \textbf{0.63} & \textbf{1.02}  & \textbf{1.74}  \\
        & Ours w/o Art.	& 0.37 &	0.56&	1.89 & 3.93	 \\
    \midrule
\multirow{2}{*}{HO3D} &     Ours & \textbf{ 0.33} &	\textbf{0.58}	& \textbf{0.93} & \textbf{4.77} \\
    &  Ours w/o Art.	& 0.27 &	0.48 &	1.18 & 6.30 \\
    \midrule
\multirow{2}{*}{MOW}   & Ours &	\textbf{0.13} &	\textbf{0.24} &	\textbf{23.1} & 19.4 \\
        & Ours w/o Art. &	0.10 & 	0.19	&	29.0 & \textbf{17.3}  \\
    \bottomrule
    \end{tabular}

    \caption{\textbf{Analysis of articulation-conditioning:} we report quantitative results of object error in F-score, Chamfer distane (CD), intersection volume on 3 datasets and  compare ours with the ablation that only consider visual feature.} 
    \label{tab:wo_hand}    
    \vspace{-1.5em}
\end{table}

\noindent \textbf{Representation of hand articulation matters for generalization. } To represent articulation information, a natural alternative to our proposed articulation-aware positional encoding is to simply concatenate the query point with the MANO pose parameter $\theta_A$, \ie $\bar{h}(\mathbf x;\theta_A) = [\mathbf x, \theta_A]$. The result in Table \ref{tab:articulation} shows that 
although it performs comparably when provided with ground-truth hand pose parameters, it degrades significantly when with predicted hand pose despite that we perform jitter augmentation on hand articulation for both methods.  More interestingly, it  performs even worse than the variant without articulation-aware encoding. This indicates that the object shape overfits to  pose parameters which are constant within one example. In contrast, our articulation-aware positional encoding generalizes better.

\begin{table}[]
\footnotesize
    \centering
    \begin{tabular}{l  cccc }
    \toprule
   Method  & F5 $\uparrow$ & F10 $\uparrow$  & CD $\downarrow$ & Vol$\downarrow$ \\  
     \midrule
        Art.-aware PE*	& \textbf{0.49} &	\textbf{0.70} &	\textbf{0.92} 	& 1.73 \\
        Pose param.*  &	0.46 &	0.66	& 1.25	 &	\textbf{1.44} \\
    \midrule
      Art.-aware PE & \textbf{0.42} & \textbf{0.63} & \textbf{1.02}  & \textbf{1.74} \\
    Pose param.	& 0.23 & 0.42 &	1.82  &	2.57 \\
    W/o art.	& 0.37 &	0.56&	1.89&	3.93    \\    
    \bottomrule
    \end{tabular}
    \caption{\textbf{Analysis of articulation-aware encoding:}  We compare different ways to incorporate hand articulation: articulation-aware positional encoding and pose parameters. Star indicates reconstruct object shape given ground truth hand articulation. }
    \label{tab:articulation}
\end{table}

\begin{table}[]
\footnotesize
    \centering
    \setlength\tabcolsep{4pt} 
    {\renewcommand{\arraystretch}{0.53}
    \vspace{-1em}
    \begin{tabular}{ll ccc ccc}
    \toprule
 & Noise & \multicolumn{3}{c}{ObMan} &  \multicolumn{3}{c}{HO3D} \\
 &   Level  & F5 $\uparrow$ & F10 $\uparrow$  & CD $\downarrow$ & F5 $\uparrow$ & F10 $\uparrow$  & CD $\downarrow$  \\  
    \midrule
 \multirow{3}{*}{Gaussian}    & 50\% $\sigma$ & 0.40 &	0.63 &	1.01 & 0.28 &	0.50 &	1.51 \\ 
    &   100\% $\sigma$ & 0.31	& 0.53 &	1.40 & 0.25 &	0.46 &	1.68 \\
    & 150\% $\sigma$ & 0.24 &	0.42 &	1.94 & 0.22 & 	0.42 &	1.93\\
    \midrule
\multirow{3}{*}{Prediction} &    50\% & 0.46	& 0.67 & 0.96 & 0.29 &	0.52 &	1.48\\
    & 100\% * & 0.42 & 0.63  & 1.02 & 0.28	& 0.50 &	1.53\\
    & 200\% & 0.35 &	0.56 & 1.28 & 0.26 &	0.47 &	1.67\\
    \midrule 
\multirow{2}{*}{Baselines} &   HO & 0.23	& 0.56	& 0.64 & 0.11 &	0.22 &	4.19  \\
    & GF & 0.30 & 0.51 & 1.39 & 0.12 &	0.24	& 4.96 \\    
    \midrule
GT &  0\% 	& 0.49 &	0.70 &	0.92 & 0.30 &	0.53 &	1.46 \\
    \bottomrule
    \end{tabular}
    }
    \caption{Error analysis against hand pose noise. $\sigma$ is the average prediction error. * marks our unablated method.  }
    \label{tab:error}
    \vspace{-3mm}
\end{table}

\begin{figure}
    \centering
    \includegraphics[width=0.98\linewidth]{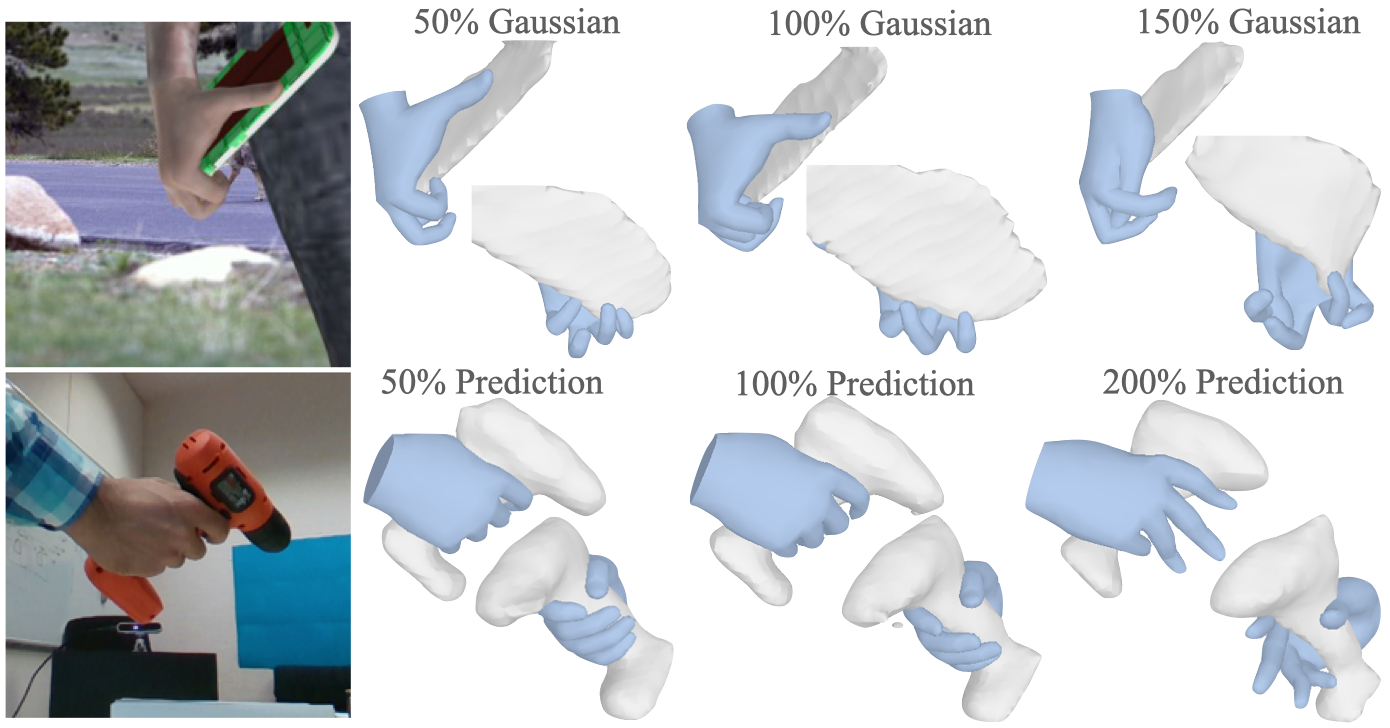}
    \vspace{-1em}
    \caption{Top: Object reconstruction given hand pose corrupted by Gaussian on Obman dataset. Bottom: Object reconstruction given hand pose corrupted by prediction error on HO3D dataset. }
    \vspace{-1em}
    \label{fig:error}
\end{figure}

\noindent\textbf{Robustness against hand prediction quality. } We use hand poses corrupted by different levels of noise, either from Gaussian or more structured prediction noise. For the latter, we linearly interpolate (and even extrapolate) the true poses and off-the-shelf predictions. Our method still outperform baselines even when the predicted hand pose is \textit{with twice more error} (Tab~\ref{tab:error} and Fig~\ref{fig:error}). 

\begin{table}[]
    \centering\footnotesize
    \begin{tabular}{l ccccc}
    \toprule
      &  F-5$\uparrow$ & CD$\downarrow$ & Vol$\downarrow$ & Sim$\downarrow$ & EPE$\downarrow$ \\
    \midrule
ours w/o rf & 0.17 &1.02  & 1.74 & 3.32 & 8.9 \\
ours w rf & 0.17 & \textbf{1.00}  & \textbf{1.28} & \textbf{3.00} & \textbf{8.7} \\ 
    \midrule
ours w GT pose & 0.20 & 0.92 & 1.73& 2.44 & --  \\
    \bottomrule
    \end{tabular}
    \vspace{-1mm}
    \caption{\textbf{Test-time refinement.} We report object error, intersection volume, simulation displacement and hand error before and after test-time refinement.} 
    \vspace{-1em}
    \label{tab:refine}
\end{table}

\begin{figure}
    \centering
    \includegraphics[width=\linewidth]{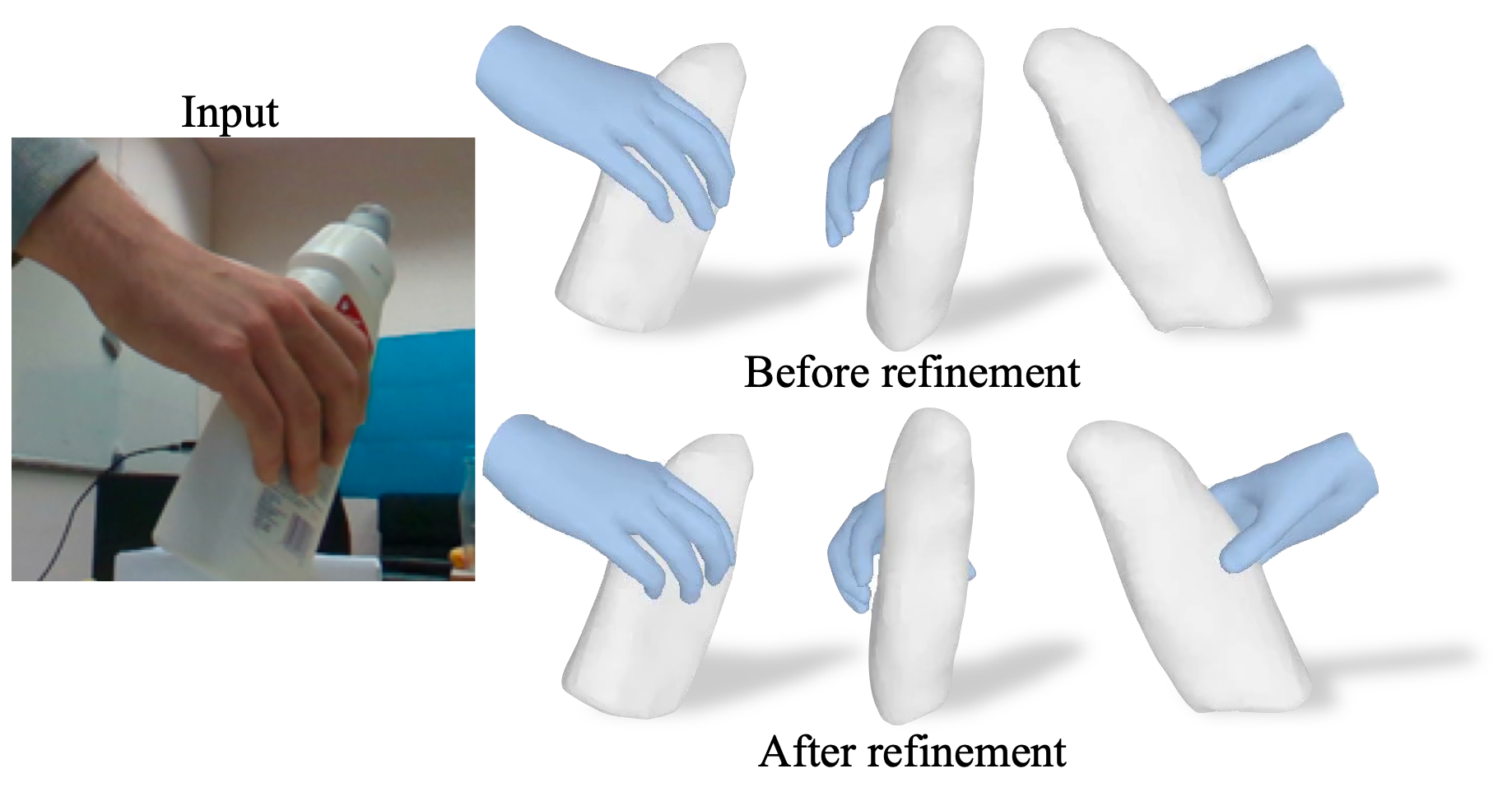}
    \caption{Visualizing hand-object reconstruction before and after test-time refinement in the image frame and from two novel views. }
    \label{fig:refine}
    \vspace{-3mm}
\end{figure}
\noindent \textbf{Test-time refinement improves hand pose.}
The object reconstruction above is obtained by direct feed-forward prediction. We then show that our articulation-conditioned object shape can in turn refine the initial hand pose estimation and improve the HOI quality. 
We report end point error (EPE in $mm$) for each joint on Obman dataset in Table \ref{tab:refine}. To evaluate HOI quality, we report intersection volume along with simulation displacement of the object. We follow prior works  \cite{hasson19_obman,tzionas2016capturing,karunratanakul2020grasping} to pass the  HOI reconstruction to a simulator and report how much the object slips from hand after running simulation for a fixed amount of time. 

As shown in Table \ref{tab:refine},  both object and hand reconstruction improve after test-time refinement. The  simulation displacement of the object drops with less intersection region.  When ground truth hand articulation is provided, the object error and simulation displacement continue to improve. 
Figure \ref{fig:refine} visualizes one example before and after refinement. Four finger tips are attracted to object surface while the thumb is pushed out of the object.




\section{Conclusion}

In this paper, we propose a method to infer implicit 3D shape of generic objects in hand. 
We explicitly treat predicted hand pose as a cue for object inference via an articulation-aware positional encoding. We have shown that this complements visual cues,  especially when the hand-held object is occluded.  While the results are encouraging, there are several limitations. For example, our work cannot be directly adapted to reconstructing dynamic grasps from videos where object consistency given varying articulation is required. Additionally, we require 3D ground-truth for training and it would be interesting to extend it with differentiable rendering techniques. Despite these challenges, we believe that our work on reconstructing hand-held generic objects takes an encouraging  step towards understanding HOI for in-the-wild videos.

\noindent\textbf{Acknowledgements.}
The authors would like to thank Zhe Cao and Ilija Radosavovic for helping set up MOW dataset.
We would also like to thank Aljosa Osep, Pedro Morgado, and Jason Zhang for their detailed feedback on manuscript.

{\small
\bibliographystyle{ieee_fullname}
\bibliography{ref}
}

\clearpage
\appendix
\section{Implementation Detail}
\subsection{Camera conversion}

The off-the-shelf system predicts a weak perspective camera with a scale factor $s$ and 2D translation $t_x, t_y$.  One can transform the point via the global hand rotation and translation and then project it via the predicted camera $s, t_x, t_y$. 
$$s T_{\theta_w} X + (t_x, t_y)$$
We found that a full perspective camera help to account for large perspective effect.  Therefore, we convert the weak perspective camera to a full perspective one by translating the final mesh by an offset $(t_x, t_y, f/s)$. In summary, we project a query point in the wrist frame to the image by 
$$
\pi_{\theta_w}(X) = K[T_{\theta_w}X + (t_x, t_y, f/s)]
$$

\subsection{Coordinate Transformation}
Our articulation embedder takes as input an articulation parameter $\theta_A$ and a point position in wrist frame $X$ to output the articulation-aware encoding $\psi = h(X;\theta_A)$. 
The encoding is a concatenation of the coordinates relative to every joint. Given the articulation parameter $\theta_A$, we run forward kinematics to derive transformation $T(\theta_A): \mathbb R ^3 \to \mathbb R ^ {45}$ that maps a point in wrist frame to each joint coordinate. 

The transformation between wrist to one joint $T_j$ is computed by forward kinematics chain. 
Consider one bone that connects joint $j$ to its child $i$ (\eg index proximal phalanx).  The transformation matrix from this joint frame to its child joint frame would be 
$$T_{ji} = \begin{pmatrix}R(\theta_j) & t_{ji}\\ 0 & 1 \end{pmatrix}$$ where $t_j$ is the bone length pre-defined in MANO models. 
Then the transformation from wrist to any joint is the product of every transformation in the kinematic chain $T_j = T_{wi}\cdot T_{ik} \cdot \dots T_{lj}$. The coordinate of the queried point relative to the joint becomes $^jX = ^jT_w ^wX$. 

\subsection{Training}
We train our model using Adam optimizer with learning rate $1e-4$ on 8 GPUs. The batch size is 64.  We train our model on ObMan for 200 epochs and finetune it on HO3D and RHOI for 50k iterations respectively. The coefficient of eikonal term is $0.1$. 

\subsection{HO3D dataset split}
HO-3D\cite{hampali2020honnotate} is a real-world video dataset consisting of 103k annotated images capturing 10 subjects interacting with 10 common YCB objects~\cite{alli2015TheYO}. 
The original train-test splits are created by partitioning the interaction sequences. 
Sequences in the original test set  involve only 4 objects of which three appear in train set (bleach cleanser, mustard bottle, meat can) and all of them are cuboidal shape. 
To test on more non-trivial shapes like power drill, we create a custom split by holding out one video sequence per object as test set. 
We list our sequences for train and test set in Table \ref{tab:ho3d_split}. 

\begin{table*}[]
    \centering\small
    \begin{tabular}{l ll}
Objects & Test Sequences & Train Sequence\\
\midrule
010\_potted\_meat\_can &GPMF10& MPM14, GPMF13, MPM12, GPMF12, MPM11, GPMF11, MPM13, MPM10, GPMF14  \\
021\_bleach\_cleanser &ABF10 & SB11, SB12, ABF11, ABF13, SB10, ABF12, ABF14, SB13, SB14  \\
019\_pitcher\_base &AP10 & AP11, AP14, AP13, AP12  \\
003\_cracker\_box & MC1 & MC2, MC6, MC5, MC4  \\
006\_mustard\_bottle &SM1 & SM5, SM2, SM4, SM3  \\
004\_sugar\_box & SS1 & ShSu12, SiS1, SS2, ShSu14, ShSu13, SS3, ShSu10  \\
035\_power\_drill & MDF10 & MDF12, MDF14, MDF11, ND2, MDF13  \\
011\_banana & BB10 & BB12, SiBF10, SiBF14, SiBF11, SiBF12, BB13, BB11, SiBF13, BB14  \\
037\_scissors & GSF10 & GSF13, GSF12, GSF14, GSF11  \\
025\_mug & SMu1 & SMu41, SMu42, SMu40  \\

    \end{tabular}
    \caption{Our customized split on HO3D dataset.}
    \label{tab:ho3d_split}
\end{table*}

\section{Qualitative Results}
We provide more qualitative results rendered in the image frame and from another view in this PDF and video results when moving camera around the object in the zipped website. 

\begin{figure*}
    \centering
    \includegraphics[width=\linewidth]{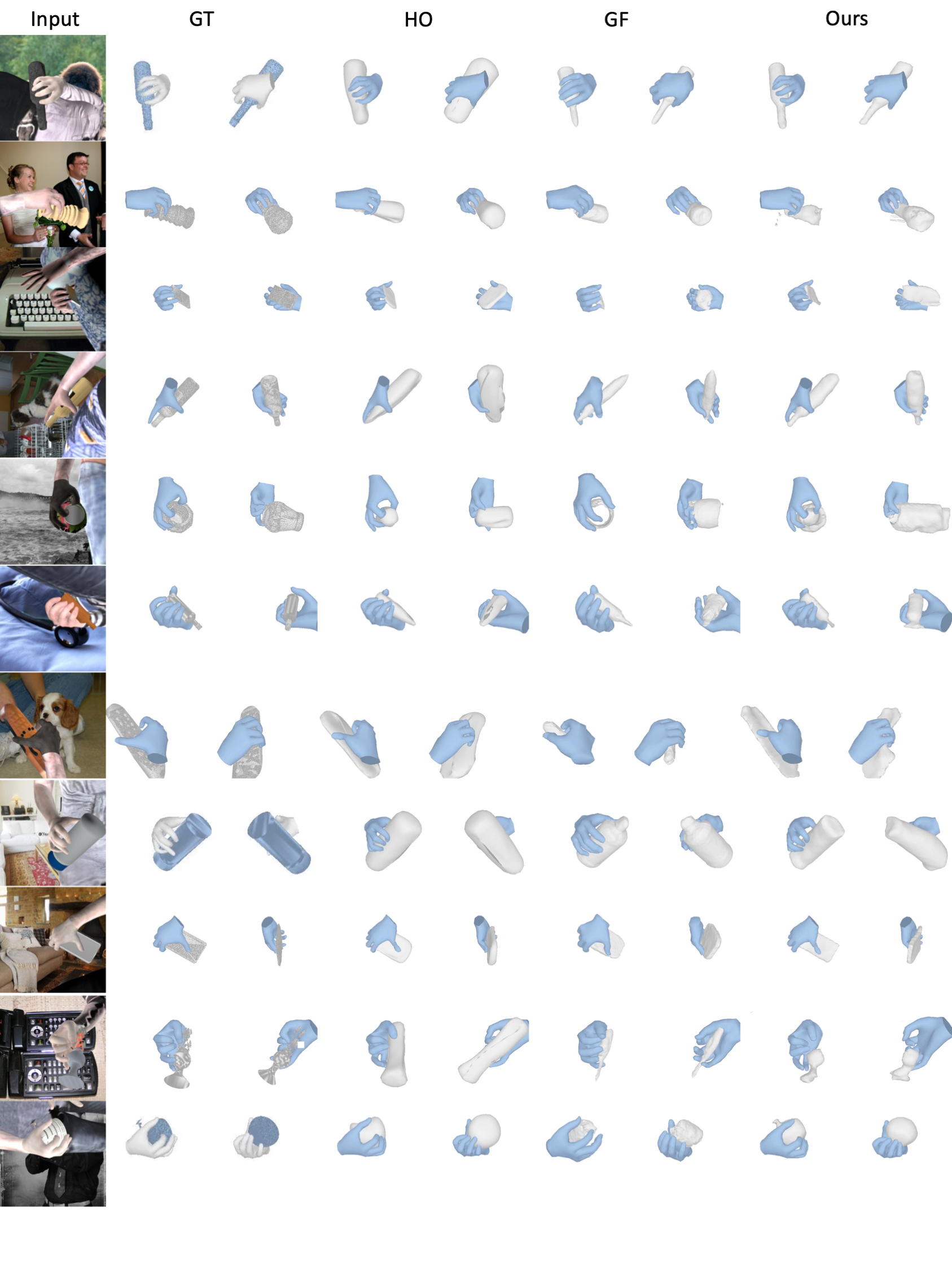}
    \caption{Visualizing reconstruction from our method and two baselines\cite{hasson19_obman, karunratanakul2020grasping} on ObMan dataset from the image frame and a novel view.}
    \label{fig:obman}
\end{figure*}

Figure \ref{fig:obman} visualizes reconstruction from our method and two baselines \cite{hasson19_obman, karunratanakul2020grasping} on ObMan dataset from the image frame and a novel view.

Figure \ref{fig:ho3d} visualizes reconstruction from our method and two baselines \cite{hasson19_obman, karunratanakul2020grasping} on HO3D dataset from the image frame and a novel view.

\begin{figure*}
    \centering
    \includegraphics[width=\linewidth]{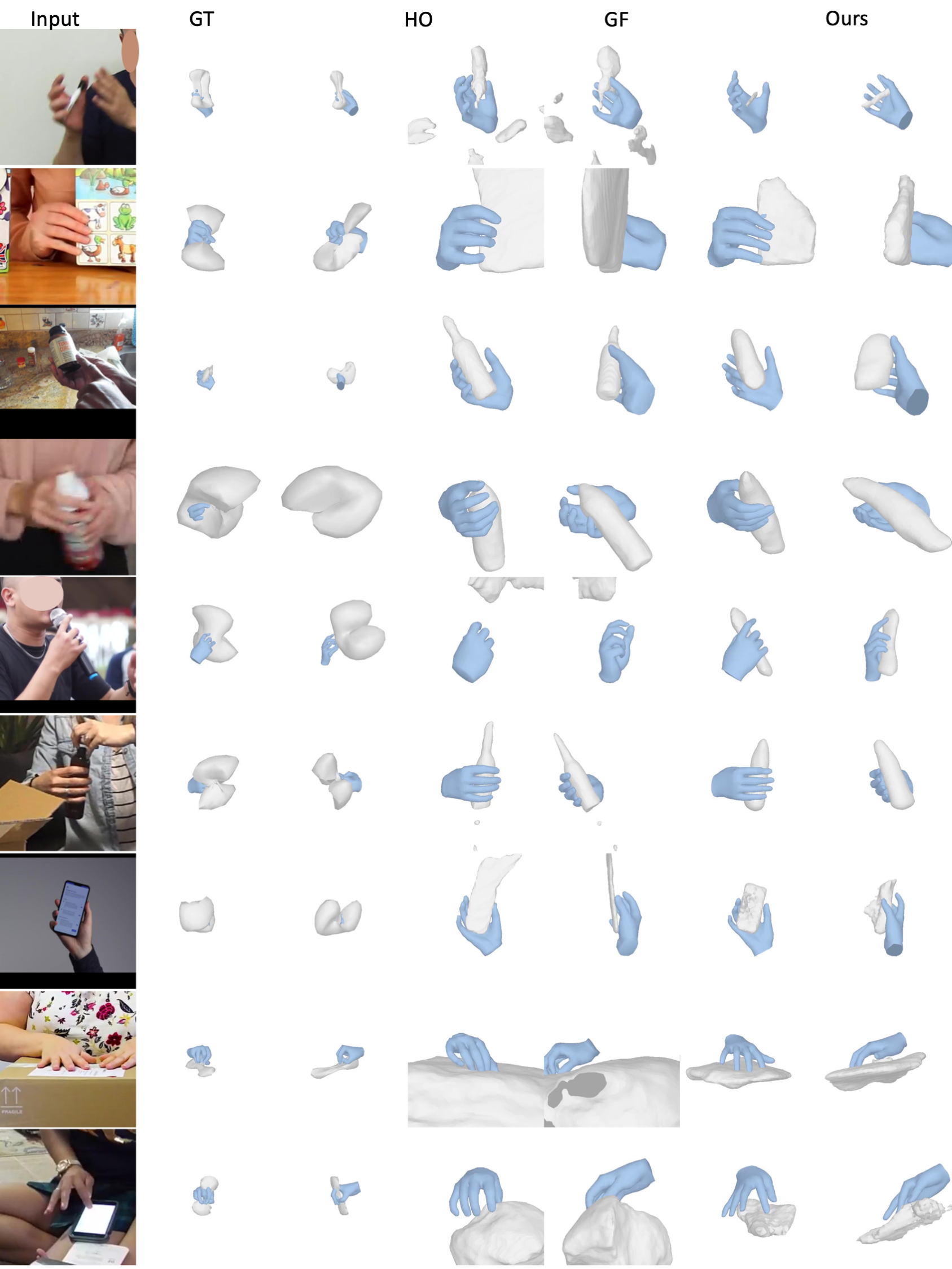}
    \caption{Visualizing reconstruction from our method and two baselines\cite{hasson19_obman, karunratanakul2020grasping} on RHOI dataset from the image frame and a novel view. }
    \label{fig:rhoi1}
\end{figure*}

\begin{figure*}
    \centering
    \includegraphics[width=\linewidth]{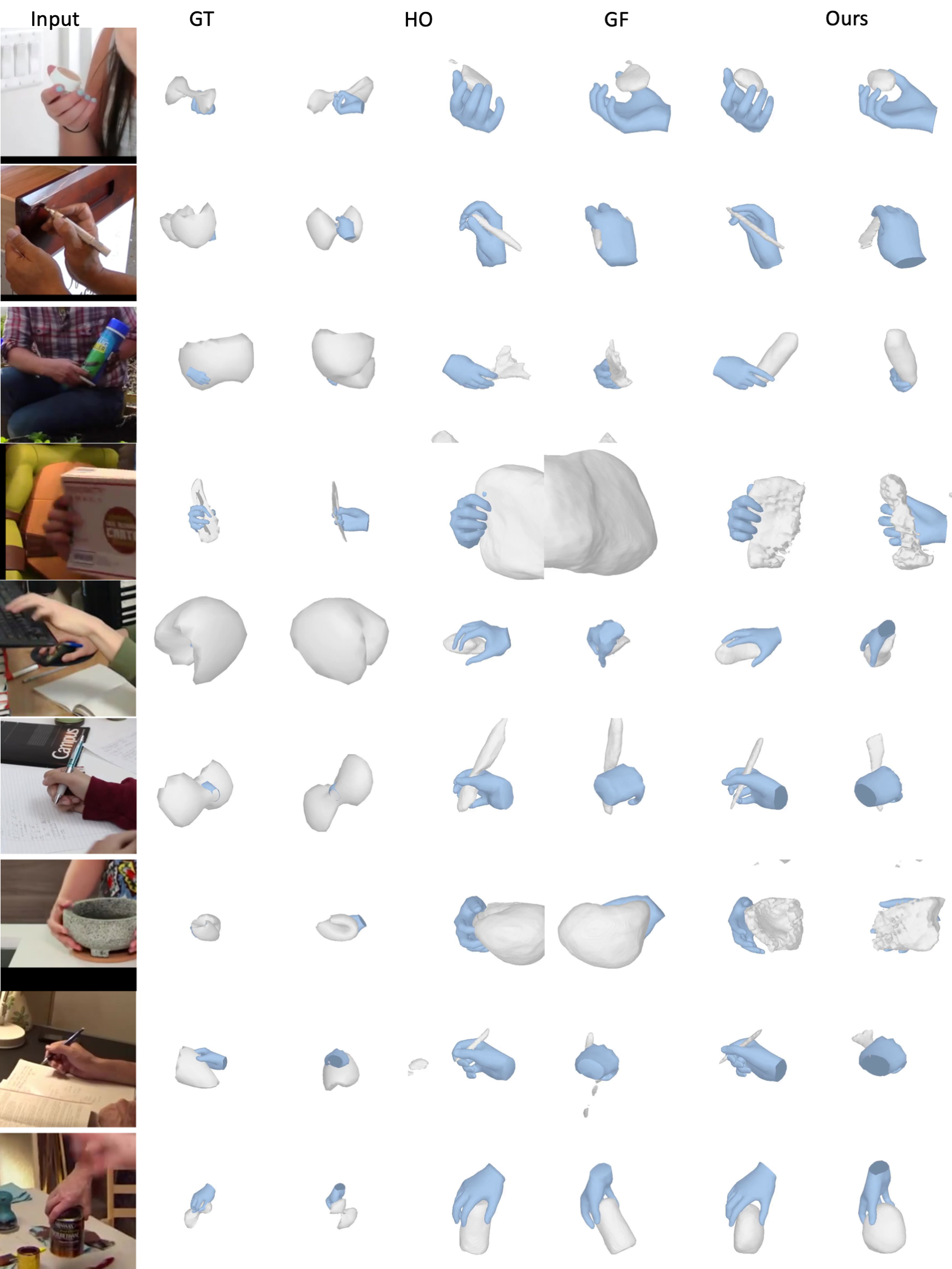}
    \caption{Visualizing reconstruction from our method and two baselines\cite{hasson19_obman, karunratanakul2020grasping} on RHOI dataset from the image frame and a novel view. }
    \label{fig:rhoi2}
\end{figure*}

Figure \ref{fig:rhoi1}, \ref{fig:rhoi2} visualizes reconstruction from our method and two baselines \cite{hasson19_obman, karunratanakul2020grasping} on RHOI dataset from the image frame and a novel view.

\begin{figure*}
    \centering
    \includegraphics[width=\linewidth]{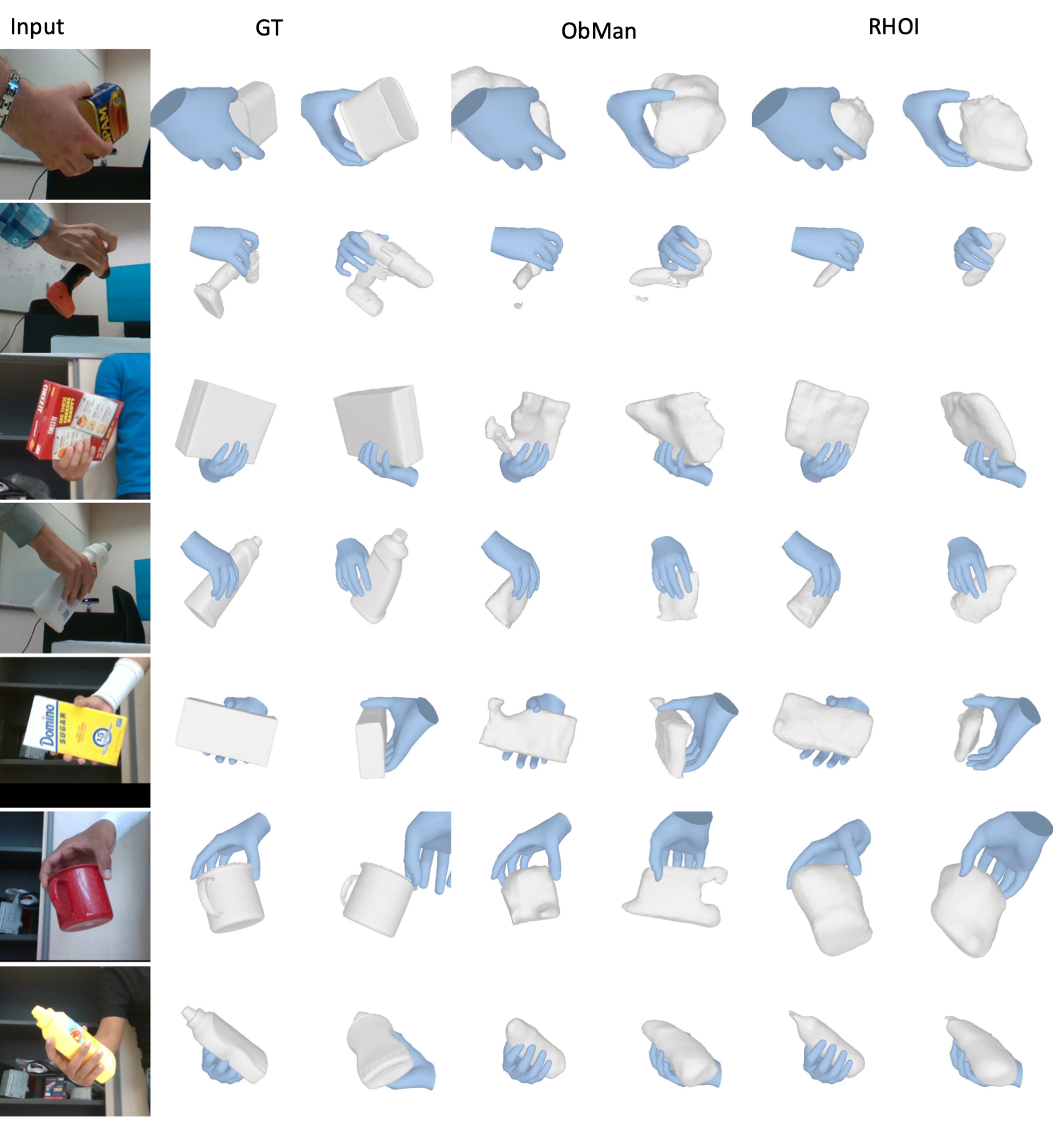}
    \caption{\textbf{Cross-dataset generalization: } we show quantitative results on HO3D for models pretrained on ObMan and RHOI. }
    \label{fig:zero}
\end{figure*}

Figure \ref{fig:zero} visualizes reconstruction of hand-held object with or without explicitly considering  hand pose on ObMan, HO3D and RHOI.   

\begin{figure*}
    \centering
    \includegraphics[width=\linewidth]{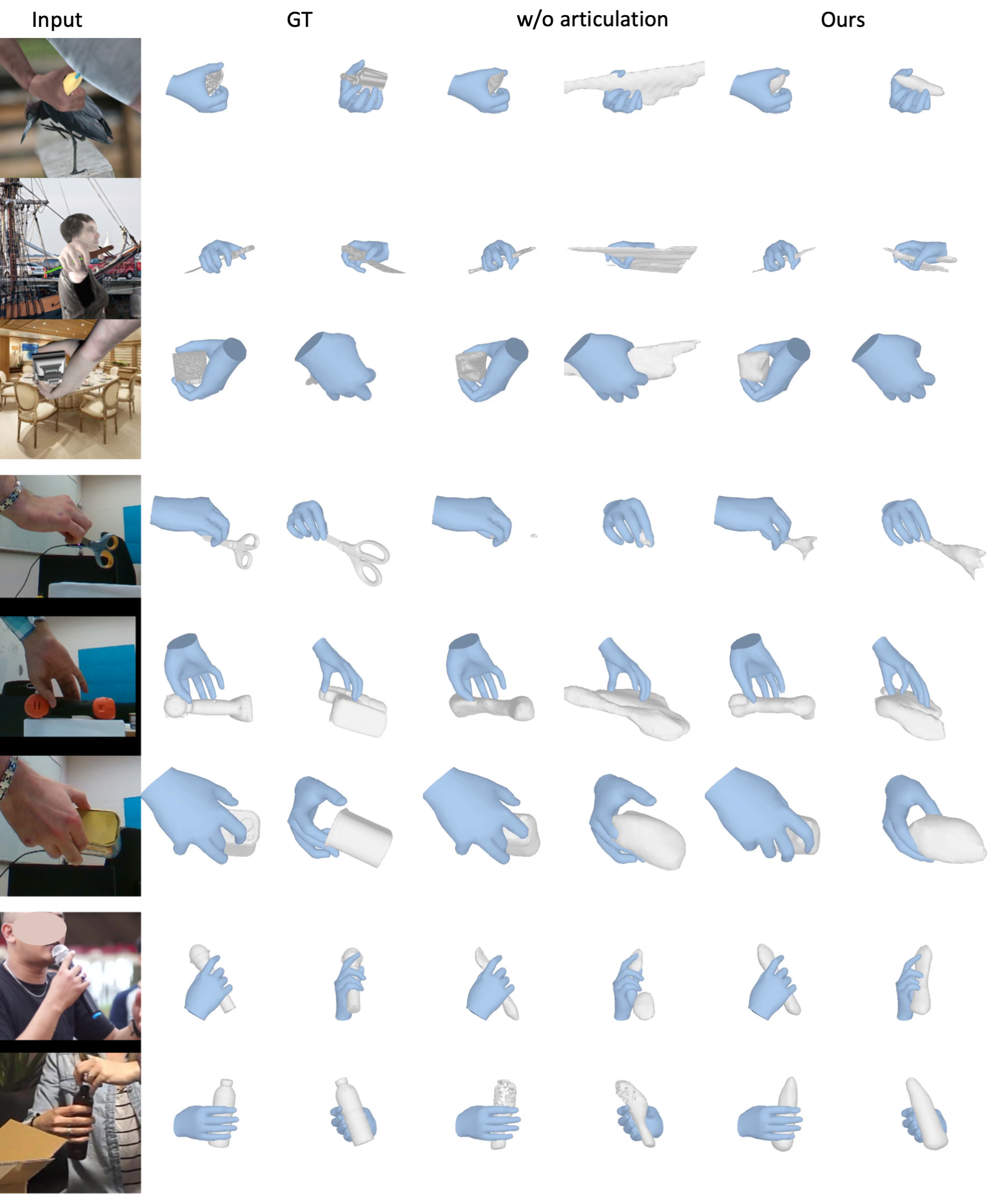}
    \caption{Visualizing reconstruction of hand-held object with or without explicitly considering  hand pose on ObMan, HO3D and RHOI.   }
    \label{fig:wo_art}
\end{figure*}

Figure \ref{fig:wo_art} visualizes reconstruction of hand-held object from our models that only trained on ObMan and RHOI datasets.

Figure \ref{fig:refine} visualizes hand-object reconstruction before and after test-time refinement in the image frame and from two novel views.  

\begin{figure*}
    \centering
    \includegraphics[width=\linewidth]{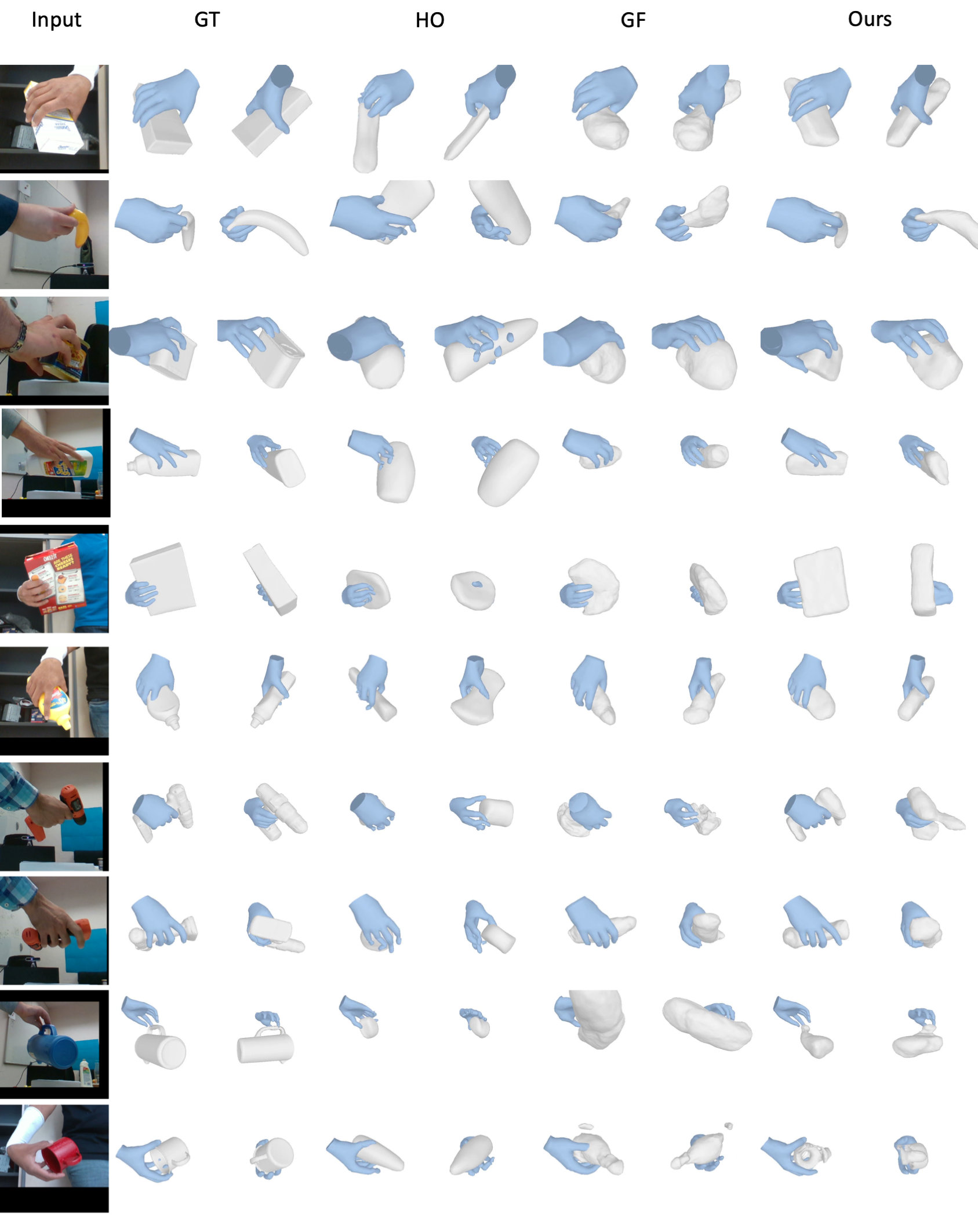}
    \caption{Visualizing reconstruction from our method and two baselines\cite{hasson19_obman, karunratanakul2020grasping} on HO3D dataset from the image frame and a novel view. }
    \label{fig:ho3d}
\end{figure*}

\begin{figure*}
    \centering
    \includegraphics[width=\linewidth]{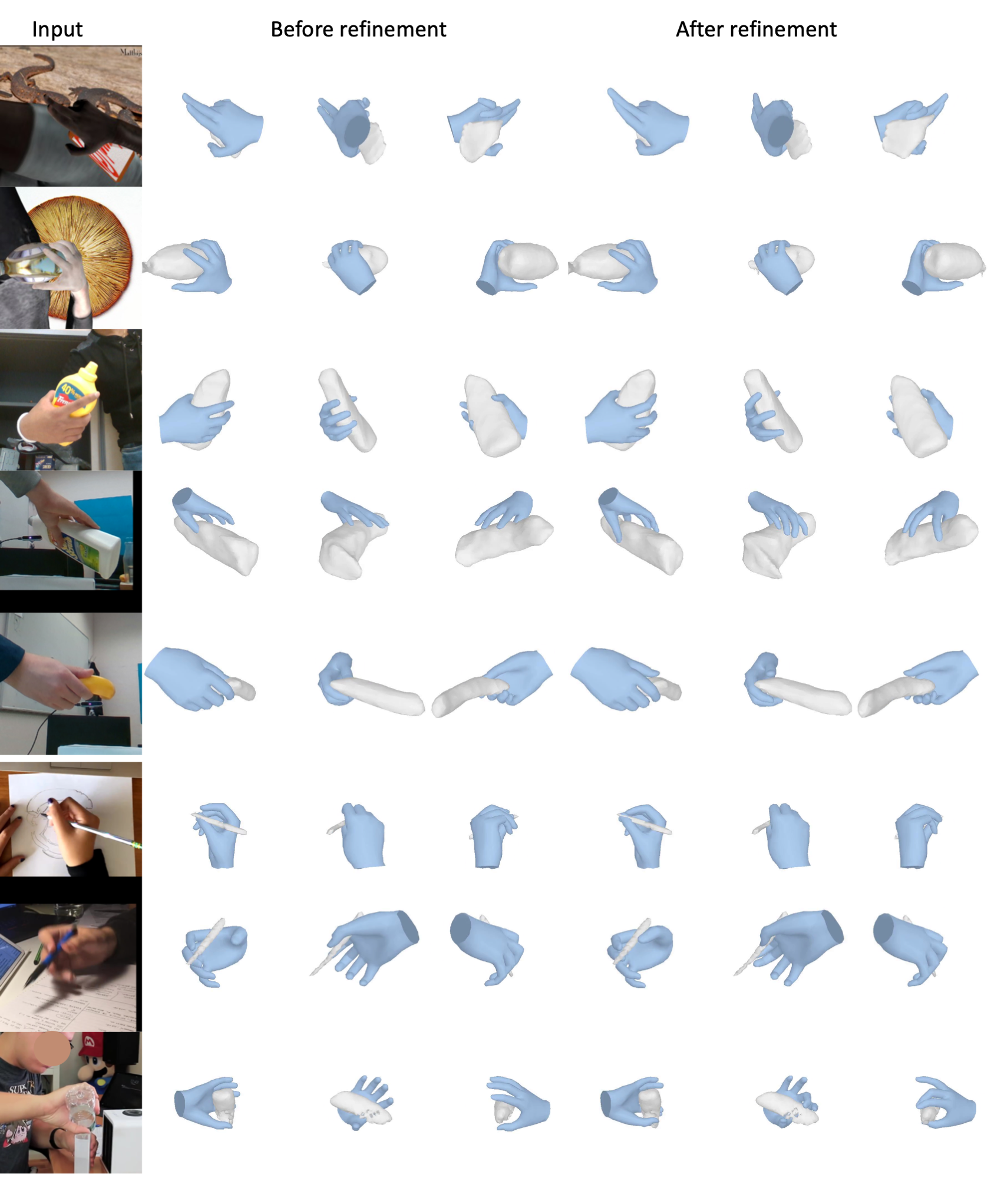}
    \caption{Visualizing hand-object reconstruction before and after test-time refinement in the image frame and from two novel views.  }
    \label{fig:refine}
\end{figure*}

\end{document}